\documentclass[runningheads]{llncs}
\usepackage[T1]{fontenc}
\usepackage{hyperref}
\usepackage{graphicx}
\usepackage[skip=5pt]{caption}
\usepackage{color}

\usepackage{float}          
\usepackage{makecell}       
\usepackage{booktabs}       
\usepackage{array}          
\usepackage{graphicx}       
\usepackage{amsmath}        
\usepackage[skip=0.5\baselineskip]{caption}
\usepackage{subcaption}
\usepackage{cite} 
\newcolumntype{T}[1]{>{\centering\arraybackslash}p{#1}}
\usepackage{multicol}
\begin{document}
\title{Enhancing Corpus Callosum Segmentation in Fetal MRI via Pathology-Informed Domain Randomization}
\titlerunning{Enhancing Corpus Callosum Segmentation in Fetal MRI}
%

\author{Marina Grifell i Plana\inst{1, 2}\textsuperscript{*}\and
Vladyslav Zalevskyi\inst{2,3}\textsuperscript{*} \and
Léa Schmidt\inst{2} \and
Yvan Gomez\inst{4,5} \and
Thomas Sanchez\inst{2,3} \and
Vincent Dunet\inst{2} \and
Mériam Koob\inst{2} \and
Vanessa Siffredi\inst{2,3} \and
Meritxell Bach Cuadra\inst{2,3}
}
\authorrunning{Grifell i Plana et al.}
%
\institute{Univeristat Politècnica de Catalunya, Barcelona, Spain \and
Department of Radiology, Lausanne University Hospital and University of Lausanne (UNIL), Lausanne, Switzerland  \and
CIBM Center for Biomedical Imaging, Lausanne, Switzerland \and
BCNatal Fetal Medicine Research Center (Hospital Clínic and Hospital Sant Joan de Déu), Universitat de Barcelona, Barcelona, Spain \and 
Department Woman-Mother-Child, CHUV, Lausanne, Switzerland\\
\textsuperscript{*}These authors contributed equally to this work. Corresponding author email: \email{vladyslav.zalevskyi@unil.ch}}

\maketitle            

\begin{abstract}
Accurate fetal brain segmentation is crucial for extracting biomarkers and assessing neurodevelopment, especially in conditions such as corpus callosum dysgenesis (CCD), which can induce drastic anatomical changes. However, the rarity of CCD severely limits annotated data, hindering the generalization of deep learning models. To address this, we propose a pathology-informed domain randomization strategy that embeds prior knowledge of CCD manifestations into a synthetic data generation pipeline. By simulating diverse brain alterations from healthy data alone, our approach enables robust segmentation without requiring pathological annotations. 

We validate our method on a cohort comprising 248 healthy fetuses, 26 with CCD, and 47 with other brain pathologies, achieving substantial improvements on CCD cases while maintaining performance on both healthy fetuses and those with other pathologies. From the predicted segmentations, we derive clinically relevant biomarkers, such as corpus callosum length (LCC) and volume, and show their utility in distinguishing CCD subtypes. Our pathology-informed augmentation reduces the LCC estimation error from 1.89 mm to 0.80 mm in healthy cases and from 10.9 mm to 0.7 mm in CCD cases. Beyond these quantitative gains, our approach yields segmentations with improved topological consistency relative to available ground truth, enabling more reliable shape-based analyses. Overall, this work demonstrates that incorporating domain-specific anatomical priors into synthetic data pipelines can effectively mitigate data scarcity and enhance analysis of rare but clinically significant malformations.

\end{abstract}

\section{Introduction}
\label{sec:intro}
The \textbf{corpus callosum} (CC) is the brain’s largest commissural fiber tract, connecting the left and right cerebral hemispheres and facilitating interhemispheric communication. Malformations of the CC have been associated with a broad spectrum of cognitive, motor, and behavioral outcomes \cite{achiron2001development}. Among congenital brain abnormalities, \textbf{corpus callosum dysgenesis} (CCD) is one of the most common, affecting approximately 0.3\%–0.7\% of the general population, and up to 2–3\% of individuals with neurodevelopmental disorders \cite{Mahallati2021}. Diagnosing CCD prenatally is challenging due to its anatomical variability and frequent association with other brain anomalies or genetic syndromes \cite{Mahallati2021}. While ultrasound (US) remains the primary screening tool for CCD, \textbf{fetal MRI can provide more information}, particularly in assessing associated abnormalities and cortical malformations \cite{Moradi2022}. When combined with super-resolution reconstruction, MRI can produce high-quality 3D volumes that reveal malformations often missed by US, providing crucial information for accurate diagnosis and prognosis \cite{Moradi2022}.

In this context, \textbf{automated segmentation of fetal brain structures} on MR images has become increasingly valuable. It enables rapid, reproducible analysis from \textit{in utero} MRI, allowing extraction of quantitative biomarkers such as shape, length, and volume, which can support early diagnosis, guide clinical decisions, and inform neurodevelopmental outcome predictions \cite{miller2016consequences, gaudfernau2021analysis}. However, \textbf{few publicly available fetal MRI datasets} exist, and those that exist are largely composed of healthy cases. For example, dHCP \cite{price2019developing_fetdHCP} includes only healthy subjects, while FeTA \cite{Payette2021b, fidon2021distributionally} provides just eight annotated cases with CCD \cite{Ciceri2024}. Such data imbalance towards healthy anatomy limits the generalization of deep learning models, which often fail on pathological cases when trained primarily on healthy data \cite{fidon2021distributionally}. 

As a result, pathology simulation has become a major focus in the field as a means to overcome data scarcity and improve model robustness. Existing pathology-informed augmentation strategies can be broadly categorized into \textbf{patch-based fusion} and \textbf{learning-based generative models}. Patch-based methods, such as CarveMix \cite{Zhang2021} and PEPSI \cite{Liu2024pepsi}, rely on lesion datasets to transplant annotated abnormalities into healthy scans via cut-and-paste operations. In contrast, learning-based approaches model pathology representations directly, using generative frameworks such as intensity- and texture-driven simulators \cite{Basaran2022}, conditional Generative Adversarial Networks (GANs) \cite{Tudosiu2024}, or diffusion models for anomaly randomization and pseudo-healthy synthesis \cite{agulia_diffcondpath_2025, Liu2025una}.  While promising, these methods \textbf{require annotated pathological data}, either for lesion sampling or representation learning, making them difficult to apply in rare disease settings.  

To reduce reliance on such annotations, some recent efforts have explored \textbf{domain randomization} (DR) as an alternative strategy. Rather than modeling realistic simulations, DR focuses on variability, through controlled, label-driven transformation, such as intensity and contrast randomization, resolution changes, and artifacts simulation, to improve robustness and \textbf{generalization to unseen domains} \cite{Billot2023, zalevskyi_drifts2024}. Shang et al.~\cite{shang2025towards} expanded the DR framework of \texttt{SynthSeg} by introducing pathology-informed label-space deformations to simulate ventriculomegaly and spina bifida in fetal cases, while using only healthy annotations. While their method improved performance on the targeted conditions, it was limited to these two pathologies and introduced minor trade-offs in healthy-case performance. Critically, \textbf{no simulation framework to date has targeted CCD}, leaving a significant gap in tools for modeling this common but underrepresented malformation.

In this work, we explore whether \textbf{prior knowledge} of CCD-related anatomical alterations, derived from medical literature and expert input, can be \textbf{embedded into a DR framework} to improve segmentation of pathological fetal brains \textit{without requiring real pathological annotations during training}. We assess whether incorporating domain-specific priors through label-space augmentations enables models trained solely on healthy data to generalize to different CCD types. Our results show that this approach substantially \textbf{improves segmentation performance on CCD cases} while preserving accuracy on healthy subjects and subjects with other pathologies. Moreover, the resulting segmentations yield anatomically consistent outputs that \textbf{support} \textbf{robust biomarker extraction}; notably, CC length derived from our automated segmentations shows \textbf{clinically meaningful differences} between CCD subgroups.

\section{Methods}
\subsection{Datasets}
This study includes a total of 370 fetal subjects sourced from both private clinical cohorts and publicly available datasets. An overview of the datasets is provided in Table~\ref{tab:datasets}; we refer readers to the original publications for full details on acquisition protocols and annotation procedures. All data were anonymized and approved by the appropriate ethics committees prior to inclusion.\footnote{
\textbf{KISPI:} Ethical Committee of the Canton of Zurich, Switzerland (Decision numbers: 2017-00885, 2016-01019, 2017-00167).
\textbf{CHUV:} Approved by the ethics committee of the Canton de Vaud (decision number CERVD 2021 00124).
\textbf{dHCP:} Approved by the UK Health Research Authority (REC: 14/LO/1169), with written parental consent.
} For each subject, we used a super-resolution reconstructed T2-weighted MRI, resampled to an isotropic resolution of 0.5×0.5×0.5mm\textsuperscript{3} and padded or cropped to a fixed size of 256×256×256 voxels. For the KISPI, dHCP, and STA datasets, annotations were harmonized into a consistent 8-class labeling scheme: cerebrospinal fluid (CSF), grey matter (GM), white matter (WM), ventricles (VM), cerebellum (CBM), deep grey matter (SGM), brainstem (BSM), and CC. Details of the harmonization procedure are provided in the Supplementary Section S1.

\noindent\textit{Training data.} Our segmentation models were trained exclusively on healthy subjects from the \textbf{KISPI} and \textbf{STA} datasets (49 cases total). These datasets provided high-quality segmentations, including the CC.

\noindent\textit{Healthy test data.} For evaluation on healthy brains, we used the \textbf{dHCP} dataset (248 subjects). This dataset was used exclusively for testing, as the CC annotations were found to be inconsistent and unsuitable for training, as discussed in Section~\ref{sec:pathresults}. Moreover, dHCP serves as an out-of-domain evaluation set to assess the generalization capabilities of our models.

\noindent\textit{CCD test data.} To evaluate segmentation performance on CCD cases, we curated a new dataset, the \textbf{CHUV-CC dataset}, consisting of 26 clinical subjects diagnosed with either complete agenesis (CCA; 12 cases, GA 23–35 weeks) or partial agenesis (pCCA; 14 cases, GA 22–32 weeks). In the pCCA cohort, we have 4 cases of hypoplasia with dysplasia (2 with a kinked morphology, 2 with a stripe-like morphology), 9 cases of hypoplasia (4 without dysplasia and 5 with an anterior remnant change) and 1 case of isolated dysplasia. Each case was reconstructed using one of three SR algorithms—NiftyMIC \cite{Ebner2020}, SVRTK \cite{Uus2023}, or NeSVoR \cite{Xu2023}—selected based on reconstruction quality. Manual segmentations of the corpus callosum were produced by an obstetrician, with more than five years of experience in fetal MRI, using ITK-Snap \cite{itksnap_py06nimg}, following the ALBERT annotation protocol \cite{Gousias2012}. Due to the high cost of manual annotation, only the corpus callosum was labeled in these cases, and only by one rater.

\noindent\textit{Other test data.} To further test generalization to unseen pathologies, we evaluated our models on 47 additional pathological cases from the \textbf{KISPI} cohort, referred to as \textbf{KISPI-Path}. These subjects exhibit a wide range of brain abnormalities, including spina bifida, ventriculomegaly, interhemispheric cysts, and heterotopia, but do not include any isolated CCD cases. This dataset allows us to assess whether models trained to simulate CCD-specific alterations maintain robustness on other fetal brain malformations.
\begin{table}[t]
\caption{Dataset properties. $N_n$ – number of neurotypical and $N_p$ – pathological subjects.}
\label{tab:datasets}
\setlength{\extrarowheight}{2pt}
\small
\resizebox{\textwidth}{!}{%
\begin{tabular}{
   p{2cm}   
   p{3.2cm} 
   p{5.2cm} 
   p{2.2cm}   
   p{2.2cm} 
   p{1.5cm} 
   p{1.2cm} 
   p{2.4cm} 
}
\toprule
\textbf{Site} & 
\textbf{Scanner} & 
\makecell[l]{\textbf{Acquisition} \\ \textbf{Parameters}} & 
\makecell[l]{\textbf{SR} \\ \textbf{Algorithm}} & 
\makecell[l]{\textbf{Resolution} \\ \textbf{($mm^3$)}} & 
\makecell[l]{\textbf{GA} \\ \textbf{(weeks)}} & 
\makecell[l]{\textbf{$N_n$} \\ \textbf{/ $N_p$}} &
\textbf{Annotations} \\
\midrule
\makecell[l]{KISPI \\ \cite{Payette2021}; \cite{Payette2021b}}
  & \makecell[l]{GE Signa \\ Discovery \\ MR450 / MR750 \\ (1.5T / 3T)}  
  & \makecell[l]{SS-FSE \\ TR/TE: 2500–3500 / 120 ms \\ 0.5×0.5×3.5 mm³} 
  & \makecell[l]{MIAL \cite{Tourbier2020} \\ IRTK \cite{KuklisovaMurgasova2012}} 
  & \makecell[l]{$0.5^3$ \\ $0.5^3$} 
  & \makecell[l]{20–34 \\ 20–34} 
  & \makecell[l]{15/23 \\ 16/24} 
  & \makecell[l]{FeTA \cite{Payette2021}+ \\ CC \cite{Fidon2021}} \\
\midrule
\makecell[l]{CHUV}
  & \makecell[l]{Siemens \\ MAGNETOM \\ Aera (1.5T)} 
  & \makecell[l]{HASTE \\ TR/TE: 1200/90 ms \\ 1.13×1.13×3 mm³} 
  & \makecell[l]{ SVRTK \cite{Uus2023} \\ NeSVoR \cite{Xu2023} \\ NiftyMIC \cite{Ebner2020}} 
  & \makecell[l]{ $0.8^3$ \\ $0.8^3$ \\ $0.8^3$ } 
  & \makecell[l]{22–35 \\ 25 \\ 25 } 
  & \makecell[l]{0/24 \\ 0/1 \\ 0/1} 
  & \makecell[l]{ Manual CC \\ Manual CC \\ Manual CC}\\
\midrule
\makecell[l]{dHCP \\ \cite{Hughes2017}}
  & \makecell[l]{Philips \\ Achieva \\ (3T)}  
  & \makecell[l]{MB‐TSE \\ TR/TE: 2265/250 ms \\ 1.1×1.1×2.2 mm³} 
  & \makecell[l]{IRTK \\ \cite{KuklisovaMurgasova2012}; \cite{Jiang2007}}
  & $0.8^3$ 
  & 21–38 
  & 248/0 
  & dHCP \cite{Makropoulos2018} \\
\midrule
\makecell[l]{STA \\ \cite{Gholipour2017}}
  & \makecell[l]{Siemens Skyra / Trio \\ (3T) \\ Philips Achieva \\ (1.5T)}  
  & \makecell[l]{SS‐FSE \\ TR/TE: 1400–2000/100–120 ms \\ 0.9–1.1×0.9–1.1×2.0 mm³} 
  & SR1 \cite{Gholipour2010} 
  & $1^3$ 
  & 21–38 
  & 18/0 
  & \makecell[l]{FeTA \cite{Payette2021}+ \\ CC \cite{Fidon2021}}\\
\bottomrule
\end{tabular}%
}
\vspace{-.4cm}
\end{table}

\subsection{Pathology-Informed Domain Randomization for CCD}
\begin{figure}[h]
\centering\includegraphics[width=0.9\linewidth]{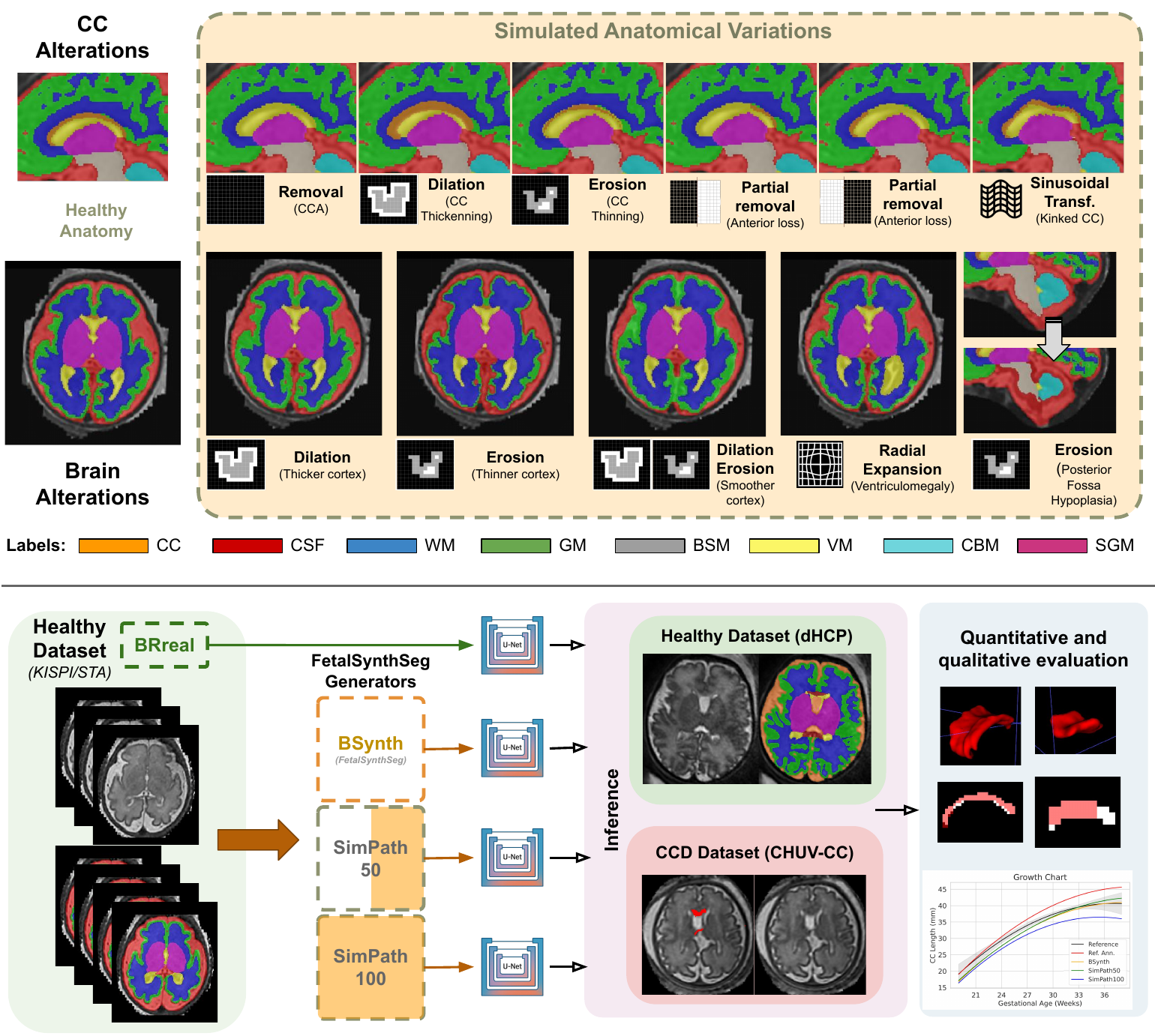}
    \caption{\footnotesize Simulation framework for CC alterations (top panel, first row) and broader brain anatomy alterations (top panel, second row), with the bottom panel showing the overall training and evaluation pipeline used in this study.}
    \label{fig:brain_alter}
    \vspace{-0.5cm}
\end{figure}

In this work, we introduce a pathology-informed extension of \texttt{FetalSynthSeg}~\cite{Zalevskyi2024}, developed to\textbf{ simulate a wide range of CC anomalies} and associated neurodevelopmental alterations, summarized in the Figure \ref{fig:brain_alter}. Our approach closely follows the original \texttt{FetalSynthSeg} pipeline but modifies its input by applying anatomical augmentations directly to the \textbf{binary label masks} used as segmentation seeds for image synthesis. These deformed segmentations serve both as the basis for generating synthetic intensity images and as training targets for supervised learning of fetal brain tissue segmentation under domain randomization. The augmentation framework is informed by clinical descriptions from Manor et al.\cite{manor2020magnetic} and refined through expert input from fetal neuroradiologists. It targets key CCD phenotypes (Figure \ref{fig:brain_alter}, top), including: \textit{complete agenesis} (simulated by replacing the CC label with the ventricles), \textit{partial agenesis}—posterior or anterior (achieved by masking out a proportional segment along the anterior–posterior axis), \textit{thinning} (implemented with binary erosion using a vertical line structuring element), \textit{thickening} (through dilation with a spherical kernel and intensity extrapolation), and \textit{kinked CC} morphology (simulated with a smooth local non-ridid sinusoidal deformation inside the CC's bounding box along the anterior–posterior axis).

To reflect the frequent co-occurrence of CCD with other central nervous system malformations, we also \textbf{simulate broader anatomical anomalies}. Cortical alterations include \textit{thickened cortex }(dilation of cortical labels constrained to gray–white matter boundaries), \textit{thinned cortex }(expansion of CSF within cortical boundaries), and \textit{smoothed cortex} (sequential dilation and erosion to reduce sulcal folding). For \textit{posterior fossa abnormalities}, we simulate brainstem and cerebellar hypoplasia (joint erosion of deep brain structures while preserving connectivity to adjacent tissue). \textit{Ventriculomegaly} is modeled using localized non-rigid radial deformation fields that expand the ventricles, with randomized lateralization, producing unilateral or bilateral presentations.

To ensure plausibility, an expert neuroradiologist reviewed the generated segmentations, both individual alterations and their combinations, to confirm broad anatomical constraints (for example, maintaining realistic ventricular dilation limits and proper continuity in simulations of pontocerebellar hypoplasia). However, the expert did not assess or correct finer‑grained details (such as subtle changes in adjacent structures like the cingulate gyrus), both because detailed segmentations of these regions were not available and because enforcing such precision would conflict with the goal of domain randomization.

 All these transformations are applied just before the intensity synthesis step and are seamlessly integrated into the \texttt{FetalSynthSeg} pipeline, which ultimately generates synthetic images used to train a \textbf{3D U-Net segmentation model} for the 8-class tissue segmentation. All augmentations are implemented in 3D and randomized over a range of severity to increase anatomical diversity. The full augmentation code, detailed explanation of the implemented augmentations and parameter ranges, as well as code and documentation, are available at the project GitHub: \href{https://github.com/Medical-Image-Analysis-Laboratory/fetalsyngen/tree/callomorph}. We also release a Docker image containing the trained model and its weights to facilitate open science efforts and ensure reproducibility and enable broader testing on external datasets \footnote{Segmentation model is available at the FetalSynthSeg DockerHub page \href{https://hub.docker.com/repository/docker/vzalevskyi/fetalsynthseg/general}{https://hub.docker.com/repository/docker/vzalevskyi/fetalsynthseg/general} under \texttt{ccpathsim} tag}.
 
\vspace{-0.2cm}

\subsection{Experimental Setup}
\looseness=-1
\noindent\textbf{Evaluated Models.}
We evaluate four segmentation models to study the effect of domain-randomized pathological augmentation. The two baselines were: \texttt{BReal}, trained exclusively on real fetal MRIs and \texttt{BSynth}, trained on synthetic images generated using the unmodified\texttt{ FetalSynthSeg} framework \cite{zalevskyi_drifts2024}. To assess the impact of incorporating simulated pathologies into the training dataset, we introduced two pathology-augmented variants: \texttt{SimPath50} and \texttt{SimPath100}, trained on the same synthetic data as \texttt{BSynth}, but with respectively 50\% and 100\% of training samples modified using our anatomical transformations. 

\noindent\textbf{Model architecture and optimization.}
In all our experiments, we use a  3D-UNet implemented with MONAI \cite{monai}, that begins with 32 channels and doubles at each upsampling stage. Convolutional layers use 3×3×3 kernels and \texttt{LeakyReLU} activations, except for the final layer, which uses softmax. Skip connections link encoder and decoder levels to preserve spatial resolution. Models were trained with the Adam optimizer (learning rate $10^{-3}$) using a hybrid Dice–Cross Entropy loss. Training stability was managed with a \texttt{ReduceLROnPlateau} scheduler (factor 0.1, patience = 10) and early stopping (patience = 100 iterations). All experiments were conducted on Nvidia RTX 6000 GPUs using PyTorch Lightning with a batch size of 1.\\
\noindent\textbf{Experiments.}
\textit{First}, we evaluate model performance on three test sets: dHCP dataset, CHUV-CC dataset, and KISPI-Path. For dHCP, we report the global average Dice score across the 7 FeTA labels, with the CC merged into white matter in both predictions and ground truth, to account for the unreliable CC annotations in this dataset, as discussed later. For CHUV-CC, we report metrics only for the CC, as it is the only annotated structure. For KISPI-Path, we report metrics for each of the 8 labels individually, as well as the global average. \textit{Second}, we assess the clinical utility of the resulting segmentations by automatically extracting CC length and evaluating its effectiveness in distinguishing between healthy, pCCA, and CCA cases. This downstream analysis serves as a proxy for the model’s suitability in supporting the extraction of clinically relevant biomarkers.\\
\noindent\textbf{Model training and architecture.}
To isolate the effect of synthetic anatomical alterations, we trained the aforementioned series of models on exclusively healthy fetal brain data from the KISPI and STA datasets, using a hybrid Dice Cross Entropy loss, for 8-class  brain tissue segmentation. Validation and testing were performed exclusively on real MRI images.  All experiments use a five-level 3D U-Net implemented in PyTorch via the MONAI framework, with exactly the same architecture and optimization parameters as in \cite{Zalevskyi2024}.\\
\noindent\textbf{Metrics and Statistical Analysis.}
We use standard segmentation evaluation metrics (described in details in the Supp. Section  S2.1: generalized Dice Score (DSC) \cite{Sudre2017}, 95th percentile Hausdorff Distance (HD95, reported in mm) \cite{Huttenlocher}, and Volume Similarity (VS) \cite{Taha2015}. For the CC label, topological correctness was additionally evaluated using the Euler Difference~\cite{Payette2025}, defined as $ED = \left|1 - (B_0 - B_1 + B_2)\right|$, where $B_0$, $B_1$, and $B_2$ are the \textit{Betti numbers} representing the number of connected components, holes, and enclosed cavities, respectively. For the CC, ground truth topology corresponds to a single connected structure with no holes or cavities, and we therefore set the ground truth $B_0 = 1$, $B_1 = 0$, and $B_2 = 0$. For clinical biomarker evaluation, we compute the deviation from normative CC length growth curves as $\Delta_{\text{Growth}} = |SC_i - RC_i|$ (mm), where $SC_i$ is the model’s predicted CC length (estimated as the distance between two furthest points of CC segmentation along the posterior-anterior axis) and $RC_i$ is normative values based on the clinical reference from \cite{Lamon2024} for a given subject at their gestational age $i$. Additional biomarker metrics include absolute CC length error ($\Delta_{\text{Length}} = |CCL_{\text{pred}} - CCL_{\text{gt}}|$). \textbf{For CCA cases,  only $\Delta_{\text{Length}}$ is computed among all metrics}, since ground truth segmentations are not available for subjects with CCA. Statistical significance is determined using the Shapiro–Wilk test for normality. When normality holds, paired Student’s t-tests are used; otherwise, Wilcoxon signed-rank tests are applied. 

\section{Results and discussion}
\subsection{Effect of Pathology-Informed Augmentations}

\begin{figure}[h]

\centering
\captionof{table}{Evaluation of the proposed pathology-informed augmentation models on the dHCP and CHUV-CC datasets. For dHCP, Dice and HD95 scores are averaged over the 7 FeTA classes, with CC being incorporated into the WM, due to CC inconsistencies in its ground truth annotations (see Figure~\ref{fig:resultsdhcp}). Statistically significant difference between the current model and BSynth is indicated by *}

\label{tab:summary_results}
\setlength{\extrarowheight}{2pt}
\small
\resizebox{\textwidth}{!}{
\begin{tabular}{
  >{\arraybackslash}m{2.7cm}
  >{\centering\arraybackslash}m{2cm}
  >{\centering\arraybackslash}m{2cm}
  >{\centering\arraybackslash}m{2cm}
  >{\centering\arraybackslash}m{2cm}
  >{\centering\arraybackslash}m{0.5cm}
  >{\arraybackslash}m{2.7cm}
  >{\centering\arraybackslash}m{2cm}
  >{\centering\arraybackslash}m{2cm}
  >{\centering\arraybackslash}m{2cm}
  >{\centering\arraybackslash}m{2cm}
}
\cline{1-5}\cline{7-11}
\textbf{CHUV-CC Metrics} & \textbf{\texttt{BReal}} & \textbf{\texttt{BSynth}} & \textbf{\texttt{SimPath50}} & \textbf{\texttt{SimPath100}}
  & & \textbf{dHCP Metrics} & \textbf{\texttt{BReal}} & \textbf{\texttt{BSynth}} & \textbf{\texttt{SimPath50}} & \textbf{\texttt{SimPath100}} \\
\cline{1-5}\cline{7-11}
DSC (CC) (\(\uparrow\))
  & 0.36* $\pm$ 0.21
  & \underline{0.50 $\pm$ 0.15}
  & \textbf{0.50 $\pm$ 0.14}
  & 0.36* $\pm$ 0.18
  & & DSC (global) (\(\uparrow\))
  & 0.73* $\pm$ 0.10
  & 0.83 $\pm$ 0.03
  & \underline{0.83 $\pm$ 0.03}
  & \textbf{0.84* $\pm$ 0.02} \\

HD95 (CC) (\(\downarrow\))
  & 3.52* $\pm$ 1.86
  & \underline{2.87 $\pm$ 1.45}
  & \textbf{2.32* $\pm$ 1.11}
  & 3.78* $\pm$ 2.12
  & & HD95 (global) (\(\downarrow\))
  & 25.04* $\pm$ 9.21
  & 8.93 $\pm$ 6.55
  & \underline{8.13 $\pm$ 6.19}
  & \textbf{5.77* $\pm$ 4.34} \\

VS (CC) (\(\uparrow\))
  & 0.73 $\pm$ 0.15
  & \underline{0.80 $\pm$ 0.12}
  & \textbf{0.81* $\pm$ 0.17}
  & 0.68 $\pm$ 0.30
  & & VS (global) (\(\uparrow\))
  & 0.82* $\pm$ 0.16
  & 0.97 $\pm$ 0.04
  & \underline{0.98* $\pm$ 0.02}
  & \textbf{0.98* $\pm$ 0.02} \\

ED (CC) (\(\downarrow\))
  & 4.00 $\pm$ 3.59
  & 2.93 $\pm$ 2.23
  & \textbf{0.93* $\pm$ 1.27}
  & \underline{1.50 $\pm$ 1.45}
  & & ED (global) (\(\downarrow\))
  & 0.56* $\pm$ 1.42
  & \textbf{0.16 $\pm$ 0.65}
  & \underline{0.17 $\pm$ 0.50}
  & 1.16* $\pm$ 1.73 \\

$\Delta_{\text{Length}}$ (CC) (\(\downarrow\))
  & 7.60 $\pm$ 6.25
  & \underline{6.46 $\pm$ 6.61}
  & \textbf{6.11 $\pm$ 6.15}
  & 7.83 $\pm$ 8.63
  & & $\Delta_{\text{Growth}}$(global)(\(\downarrow\))
  & 1.73 $\pm$ 1.69
  & \textbf{1.63 $\pm$ 1.02}
  & \underline{1.66 $\pm$ 1.08}
  & 3.12* $\pm$ 2.09 \\
  \cline{7-11}
  
nDSC (CC) (\(\uparrow\))
& 0.45* $\pm$ 0.25
  & \textbf{0.67 $\pm$ 0.13}
  & \underline{0.65 $\pm$ 0.15}
  & 0.44* $\pm$ 0.21 & & & & & & \\[0.5mm]
\cline{1-5}
\end{tabular}
}

\vspace{-0.5cm}
\includegraphics[width=1\linewidth]{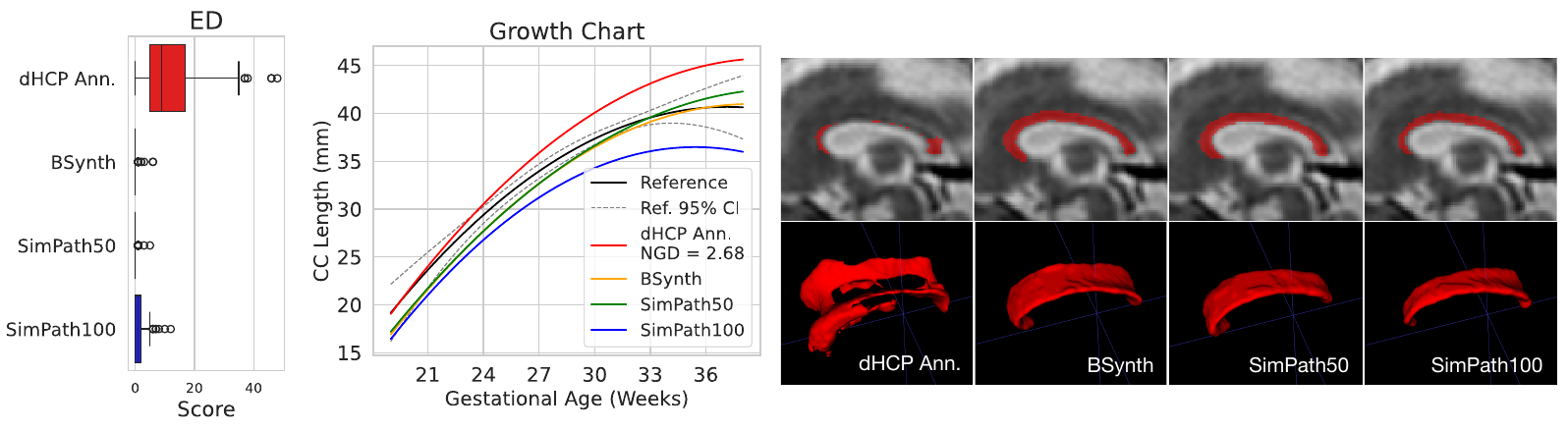}
\captionof{figure}{\textbf{Evaluation on dHCP.} \textit{Left:} Boxplot of Euler characteristic difference and CC length growth curve, comparing our models to the original dHCP annotations. \textit{Center}: CC length growth trends estimated using quadratic fits for the evaluated models. \textit{Right:} CC segmentations and meshes for representative dHCP subjects, showing improved topology and structure with our models.}
\label{fig:resultsdhcp}
\end{figure}

\begin{figure}[htbp]
  \centering
  \begin{minipage}[t]{0.44\textwidth}
    \vspace{0pt}
    \includegraphics[width=\linewidth]{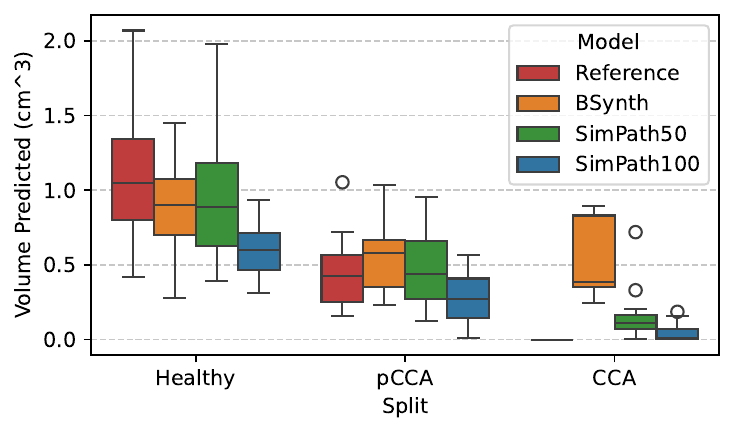}
    \label{fig:ccvolumes}
  \end{minipage}\hfill
  \begin{minipage}[t]{0.56\textwidth}
    \vspace{0pt}
    \tiny
    \begin{tabular}{lcccc}
      \toprule
      KISPI-PATH\\Metrics & BReal            & BSynth           & SimPath100       & SimPath50        \\
      \midrule
      DSC (global) (\(\uparrow\))   & 0.75\,{\scalebox{0.7}{$\pm$0.21}} & 0.75\,{\scalebox{0.7}{$\pm$0.15}} & 0.75\,{\scalebox{0.7}{$\pm$0.14}} & 0.75\,{\scalebox{0.7}{$\pm$0.14}} \\
      HD95 (global) (\(\downarrow\))  & 4.19\,{\scalebox{0.7}{$\pm$5.02}} & 3.01\,{\scalebox{0.7}{$\pm$2.27}} & 3.28\,{\scalebox{0.7}{$\pm$2.71}} & 3.07\,{\scalebox{0.7}{$\pm$2.32}} \\
      VS (global) (\(\uparrow\))   & 0.96\,{\scalebox{0.7}{$\pm$0.05}} & 0.96\,{\scalebox{0.7}{$\pm$0.04}} & 0.97\,{\scalebox{0.7}{$\pm$0.03}} & 0.97\,{\scalebox{0.7}{$\pm$0.02}} \\
      ED (global) (\(\downarrow\))   & 2.23\,{\scalebox{0.7}{$\pm$2.89}} & 1.85\,{\scalebox{0.7}{$\pm$3.75}} & 2.76\,{\scalebox{0.7}{$\pm$2.48}} & 1.31\,{\scalebox{0.7}{$\pm$2.31}} \\
      \bottomrule
    \end{tabular}
    \captionof{table}{Evaluation results on the KISPI-Path dataset, averaged across all labels.}
    \label{tab:kispipathresults}
  \end{minipage}
  \caption{CC volume distribution across healthy (dHCP) and CCD cases (left) and CC volume growth chart for dHCP cases (right).}
  \label{fig:combined-transposed}
\end{figure}

\label{sec:pathresults}
Table~\ref{tab:summary_results} summarizes the performance of models trained with and without pathology-informed augmentations. The benefits of incorporating simulated CCD alterations are particularly evident in topology-aware metrics (e.g., ED) and anatomical measures such as $\Delta_{\text{Length}}$. The \texttt{SimPath50} model consistently outperforms the baselines across all metrics, demonstrating the effectiveness of augmenting healthy data with a moderate proportion of CCD-inspired anatomical variations. In contrast, the \texttt{SimPath100} model, trained exclusively on simulated pathological anatomies, shows a decline in performance, suggesting that excessive deviation from healthy distributions may hinder generalization. Table \ref{tab:kispipathresults} as well as supplementary materials Tables  S1 and S2  further confirm that \texttt{SimPath} models maintain strong performance on non-CCD pathologies, with no observable degradation relative to the baselines.

On the dHCP dataset, models augmented with pathological simulations perform on par with the baselines, indicating that exposure to synthetic brain alteration does not negatively impact segmentation quality on healthy cases. As shown in Figure~\ref{fig:resultsdhcp}, all models trained on synthetic data using annotations from \cite{fidon2021distributionally} achieve significantly better topological consistency than the original dHCP annotations (mean ED = 0.16 for \texttt{BSynth} vs ED = 12.04 for dHCP annotations), and exhibit CC length growth trends closer to clinical reference curves ($\Delta_{\text{Growth}}$ = 1.63 mm for \texttt{BSynth}  vs 2.68 mm for dHCP annotations).

Across both CHUV-CC and dHCP test sets, the \texttt{BSynth} model consistently outperforms \texttt{BReal}, which we attribute to the fact that all evaluations are conducted OOD in terms of super-resolution method and acquisition site. The smaller performance gap observed on the KISPI-Path dataset (see Table S1) supports prior findings in \cite{zalevskyi2025_feta2024}, emphasizing that SR algorithm and acquisition site are some of the key sources of domain shift in fetal brain MRI.

\subsection{Toward Clinically Relevant Biometric Assessment}

\begin{figure}[t!]
\centering
\includegraphics[width=1\linewidth]{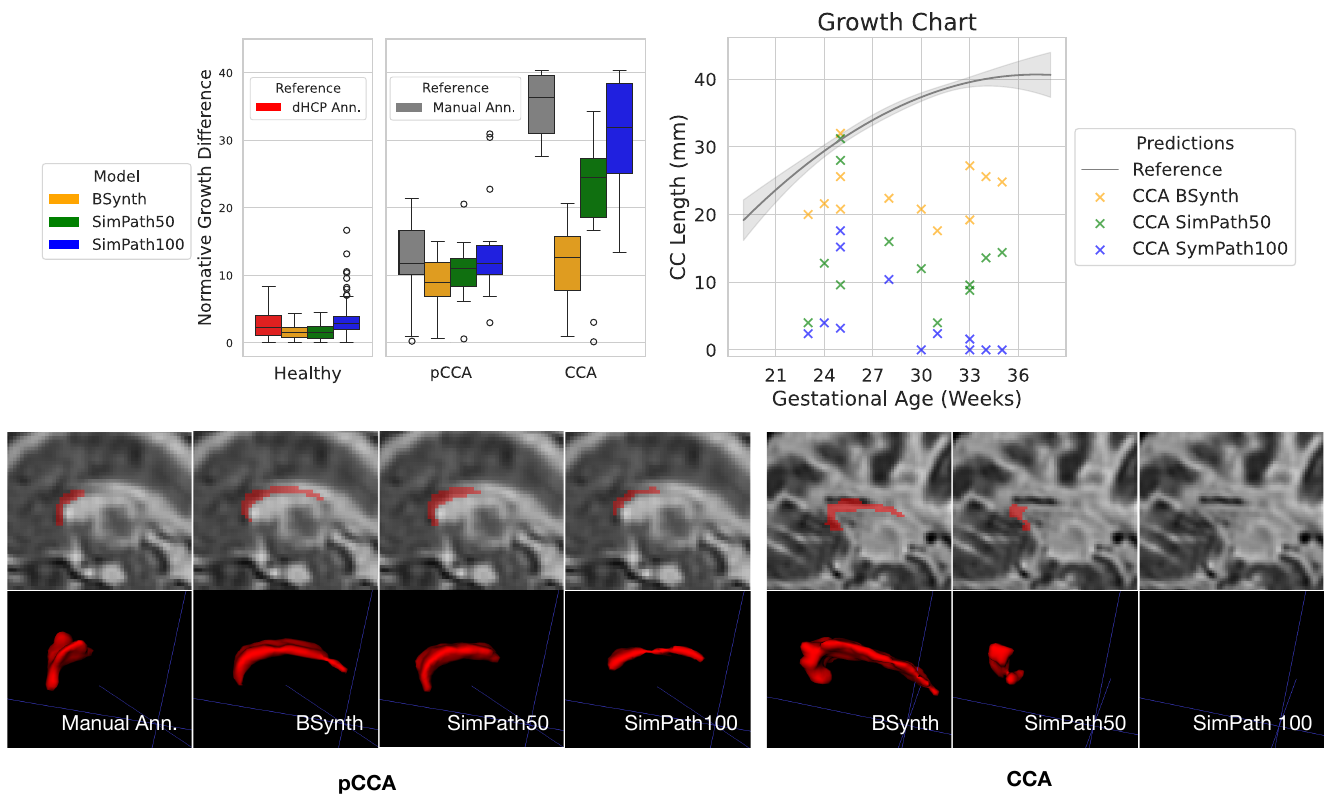}
\caption{\textbf{Evaluation on CHUV-CC}. \textit{Top left}: Deviation of predicted CC lengths from reference values across three cohorts. \textit{Top right}: CC length as a function of gestational age on CCA subjects. \textit{Bottom}: Predicted and reference segmentations and 3D CC meshes for a pCCA and CCA case.}

\vspace{-.5cm}
\label{fig:results chuv}
\end{figure}

Beyond voxel-wise accuracy, we evaluate CC length as a proxy for biometric assessment. Figures \ref{fig:combined-transposed} and ~\ref{fig:results chuv} present both qualitative and quantitative results on the CHUV-CC dataset of the CC volume and length, respectively, across healthy and CCD subjects. Figure \ref{fig:combined-transposed} shows how CC volume drastically decreases with the severity of CC dysgenesis. Furthermore, as can be seen in the Figure \ref{fig:results chuv}, the reference $\Delta_{\text{Growth}}$ increases with the severity of the pathology, indicating how CC length values differ from the expected at a given GA.  In the top row of Figure \ref{fig:results chuv}, both \texttt{SimPath50} and \texttt{SimPath100} produce CC length estimates that effectively distinguish between healthy, pCCA and CCA cases, with \texttt{SimPath100} closely matching the separation achieved by the reference annotations. In contrast, the baseline model struggles to consistently differentiate between pCCA and CCA. These findings indicate that incorporating pathology-informed augmentations enhances the model's ability to capture clinically relevant anatomical features for downstream analysis and the more severe are the pathologies in the testing dataset the more sever augmentations should be applied for an optimal performance.


\section{Conclusions}
In this study, we explored whether synthetic data generation can improve  fetal brain segmentation in presence of pathology. By embedding anatomical priors into a domain randomization framework, we simulated diverse structural alterations associated with CCD, enabling models trained solely on healthy data to generalize effectively to rare pathological cases.

Our results show that \textbf{pathology-informed augmentations improve segmentation accuracy} in CCD cases, particularly in terms of topological correctness,\textbf{ without compromising performance on normal appearing brains and provided support for clinically meaningful measurements}: corpus callosum length, for example, reliably differentiates between healthy, partial, and complete agenesis, suggesting potential for automated severity assessment.

\begin{credits}
\subsubsection{\ackname} This research was funded by the Swiss National Science Foundation (182602 and 215641) and ERA-NET Neuron MULTI-FACT project (SNSF 31NE30 203977). We acknowledge the Leenaards and Jeantet Foundations as well as CIBM Center for Biomedical Imaging, a Swiss research center of excellence founded and supported by CHUV, UNIL, EPFL, UNIGE and HUG. This research was also supported by grants from NVIDIA through an Academic Grant Program and utilized the provided NVIDIA RTX6000 ADA GPUs. The Developing Human Connectome Project (dHCP) was funded by the European Research Council (ERC) under the European Union’s Seventh Framework Programme (FP7/2007–2013), Grant Agreement No. 319456.

\subsubsection{\discintname}
The authors have no competing interests to declare that are relevant to the content of this article.
\end{credits}

%
%
%
\newpage
\begingroup
\let\clearpage\relax
\bibliographystyle{ieeetr}
{\small \bibliography{referencesshort}}
\endgroup

\makeatletter
\setcounter{table}{0}
\renewcommand 
\thesection{S\@arabic\c@section}
\renewcommand\thetable{S\@arabic\c@table}
\renewcommand \thefigure{S\@arabic\c@figure}
\makeatother

\newpage
\section*{Supplementary Material}
\small
\renewcommand{\thetable}{S\arabic{table}}

\renewcommand{\thefigure}{S\arabic{figure}}
\setcounter{table}{0}
\setcounter{figure}{0}
\setcounter{section}{0}

\section{Annotation Harmonization}
\label{sec:annatharm}
To ensure consistent training and evaluation across datasets with differing annotation protocols, we harmonized all labels to match the 7-class FeTA Challenge scheme \cite{Payette2021}, which includes cerebrospinal fluid (CSF), gray matter (GM), white matter (WM), ventricles (VM), deep gray matter (SGM), cerebellum (CBM), and brainstem (BSM). For the \textbf{KISPI} dataset, we used annotations from Fidon et al.~\cite{Fidon2021}, which refine the original FeTA labels and add a separate corpus callosum (CC) class. In the \textbf{dHCP} dataset \cite{price2019developing_fetdHCP}, annotations were produced using the Draw-EM pipeline \cite{Makropoulos2018} and manually remapped to the FeTA classes following the approach in \cite{zalevskyi_drifts2024}, with label mappings such as (Draw-EM → FeTA): 1→1 (CSF), 2→2 (GM), 3→3 (WM), 4→0 (background), 5→4 (LV), 6→5 (CBM), 7→6 (SGM), 8→7 (BSM), and 9→3 (merged with WM). Notably, this merges the hippocampi and amygdala (label 9) with WM, and the skull (label 4) with the background. Anatomical definitions also differ slightly between the two schemes—for example, Draw-EM includes only the lateral ventricles under the “ventricles” label, while FeTA includes all four ventricles. The corpus callosum label from Draw-EM was retained as an additional class. The \textbf{STA} dataset \cite{Gholipour2017} underwent the same remapping ad dHCP and also included the CC class. As a result, all datasets were standardized to an 8-class annotation scheme, comprising the 7 FeTA tissue classes plus the corpus callosum.

\section{Pathology-informed Generation Details}
\label{sec:genparams}

\subsection{Metrics}
\label{sec:metrics}



\textbf{Generalized DSC (gDSC)}~\cite{Sudre2017} applies class-specific weights to handle imbalance:
\begin{equation*}
    gDSC = 2 \frac{\sum_c w_c \cdot |S_g^c~ \cap ~S_t^c|}{\sum_c w_c(|S_g^c| ~+~ |S_t^c|)} = \frac{2 \sum_c w_c \cdot TP_c}{\sum_c w_c(2TP_c + FP_c + FN_c)}
\end{equation*}
where $w_c = 1/(S_g^c)^2$ is the inverse squared volume of class $c$.

\textbf{Hausdorff Distance 95th Quantile (HD95)}
The Hausdorff Distance measures the largest boundary discrepancy between predicted and ground truth segmentations. To mitigate sensitivity to outliers, we use the 95th percentile version~\cite{Huttenlocher}, defined as:
\begin{equation*}
    HD_{95}(A,B) = \max(h_{95}(A,B), h_{95}(B,A))
\end{equation*}
where
\begin{equation*}
    h_{95}(A,B) = p_{95}\left( \min_{b \in B} \| a - b \| ~\big|~ a \in A \right)
\end{equation*}
Distances are computed in voxels and reported in millimeters to allow comparisons across resolutions.

\textbf{Volumetric Similarity (VS)}
VS compares the total volumes of the predicted and ground truth segmentations, regardless of spatial alignment:
\begin{equation*}
    VS = 1 - \frac{||S_t| - |S_g||}{|S_t| + |S_g|}
\end{equation*}
This is particularly useful for small structures like the corpus callosum, where spatial metrics may be unreliable due to segmentation noise.

\section{Results for KISPI-Path}
Table~\ref{tab:dice_scores_transposed} reports the Dice scores of all evaluated models on the KISPI‑Path dataset for each tissue class, including the corpus callosum, while Table \ref{tab:hausdorff_transposed} presents the corresponding HD95 scores. As shown in Table~\ref{tab:kispipathresults}, the evaluated models exhibit only minor differences in their mean Dice and HD95 metrics. However, when comparing the average Euler Distance (ED) of the KISPI‑Path ground truth annotations (11.18±12.49) with the ED scores achieved by our models (Table~\ref{tab:kispipathresults}), it becomes evident that our models produce segmentations with superior topological consistency. Among all models, SimPath50 achieves the lowest ED, highlighting the effectiveness of our augmentation pipeline in preserving anatomical topology.

Notably, when considering all metrics and tissue labels, SimPath50 consistently outperforms BSynth, which serves as a more relevant baseline since both rely on synthetic training data. These results demonstrate that our pathology‑informed augmentation strategy effectively reduces real‑synthetic domain gaps, leading to improved accuracy and robustness.

\begin{table}[h]
\vspace{-.4cm}
\begin{minipage}[c]{0.48\linewidth}
\centering
\caption{Mean Dice scores ($\uparrow$) per structure across models.}
\label{tab:dice_scores_transposed}
\scriptsize
\setlength{\tabcolsep}{4pt}
\resizebox{\linewidth}{!}{\begin{tabular}{lcccc}
\toprule
Structure & BReal           & BSynth          & SimPath50          & SimPath100        \\
\midrule
CSF       & \textbf{0.68 ± 0.34} & \underline{0.64 ± 0.33} & \underline{0.64 ± 0.33} & 0.64 ± 0.34  \\
GM        & \textbf{0.68 ± 0.22} & 0.61 ± 0.24            & \underline{0.62 ± 0.22} & 0.61 ± 0.24  \\
WM        & \textbf{0.85 ± 0.17} & 0.81 ± 0.18            & \underline{0.83 ± 0.16} & 0.81 ± 0.18  \\
VM        & \textbf{0.84 ± 0.16} & 0.78 ± 0.14            & \underline{0.80 ± 0.12} & \underline{0.80 ± 0.13} \\
CBM       & 0.66 ± 0.39          & \textbf{0.78 ± 0.20}   & 0.75 ± 0.22            & \underline{0.77 ± 0.19} \\
SGM       & \textbf{0.84 ± 0.14} & 0.79 ± 0.15            & \underline{0.80 ± 0.12} & \underline{0.80 ± 0.14} \\
BS        & 0.66 ± 0.31          & \underline{0.75 ± 0.15} & 0.73 ± 0.16            & \textbf{0.76 ± 0.12}   \\
CC        & \textbf{0.63 ± 0.15} & \underline{0.56 ± 0.16} & 0.54 ± 0.14            & 0.44 ± 0.15            \\
\bottomrule
\end{tabular}}
\end{minipage}
\hfill
\begin{minipage}[c]{0.48\linewidth}
\centering
\caption{Hausdorff distances (mm, $\downarrow$) per structure across models.}
\label{tab:hausdorff_transposed}
\scriptsize
\setlength{\tabcolsep}{4pt}
\resizebox{\linewidth}{!}{\begin{tabular}{lcccc}
\toprule
Structure & BReal           & BSynth              & SimPath50             & SimPath100           \\
\midrule
CSF       & \textbf{6.78 ± 9.51} & \underline{6.95 ± 9.23} & 7.09 ± 9.52         & 7.10 ± 9.32      \\
GM        & 2.05 ± 2.22        & 1.70 ± 1.27         & \textbf{1.61 ± 1.06} & \underline{1.65 ± 1.08} \\
WM        & 3.23 ± 4.70        & 2.28 ± 1.46         & \textbf{2.03 ± 1.17} & \underline{2.10 ± 1.28} \\
VM        & 3.66 ± 4.28        & \underline{2.29 ± 1.73} & \textbf{1.99 ± 1.09} & 2.46 ± 2.41      \\
CBM       & 4.70 ± 6.66        & \textbf{2.70 ± 2.36}   & \underline{3.32 ± 3.51} & 4.27 ± 6.14     \\
SGM       & 3.17 ± 4.89        & \underline{2.70 ± 1.30} & \textbf{2.58 ± 0.82} & 2.74 ± 1.26      \\
BS        & 6.04 ± 7.38        & \textbf{2.42 ± 1.10}   & 2.87 ± 1.42         & \underline{2.71 ± 1.94} \\
CC        & \underline{3.82 ± 5.76} & \textbf{3.04 ± 4.59} & \textbf{3.04 ± 4.26} & 5.06 ± 4.53      \\
\bottomrule
\end{tabular}}
\end{minipage}
\vspace{-.4cm}
\end{table}

\end{document}